\documentclass[10pt,journal,onecolumn]{IEEEtran}
\usepackage{color}
\usepackage{graphicx}
\usepackage{amsmath}
\usepackage{array}
\usepackage{caption}
\usepackage{subcaption}
\usepackage{amsfonts}
\usepackage{amssymb}

\begin{document}
 
\title{CNN as Guided Multi-layer RECOS Transform}

\author{C.-C. Jay Kuo \\
University of Southern California
}
 
\maketitle

\thispagestyle{empty}




\section*{\bf SCOPE}\label{sec:scope}

There is a resurging interest in developing a neural-network-based
solution to the supervised machine learning problem.  The convolutional
neural network (CNN) will be studied in this note.  To begin with, we
introduce a RECOS transform as a basic building block of CNNs.  The
``RECOS" is an acronym for ``REctified-COrrelations on a Sphere"
\cite{kuo2016understanding}. It consists of two main concepts: data
clustering on a sphere and rectification.  Then, we interpret a CNN as a
network that implements the guided multi-layer RECOS transform with
three highlights. First, we compare the traditional single-layer and
modern multi-layer signal analysis approaches and point out key
ingredients that enable the multi-layer approach. Second, we provide a
full explanation to the operating principle of CNNs. Third, we discuss
how guidance is provided by labels through backpropagation (BP) in the
training. 

\section*{\bf RELEVANCE}\label{sec:relevance}

CNNs are widely used in the computer vision field nowadays.  They offer
state-of-the-art solutions to many challenging vision and image
processing problems such as object detection, scene classification, room
layout estimation, semantic segmentation, image super-resolution, image
restoration, object tracking, etc. They are the main stream machine
learning tool for big visual data analytics.  A great amount of effort
has been devoted to the interpretability of CNNs based on various
disciplines and tools, such as approximation theory, optimization theory
and visualization techniques. We explain the CNN operating principle
using data clustering, rectification and transform, which are
familiar to researchers and engineers in the signal processing and
pattern recognition community. As compared with other studies, this
approach appears to be more direct and insightful. It is expected to
contribute to further research advancement on CNNs. 

\section*{\bf PREREQUISITES}\label{sec:prerequisites}

The prerequisites consist of basic calculus, probability and linear
algebra. Statistics and approximation techniques could also be useful
but not necessary. 

\section*{\bf PROBLEM STATEMENT}\label{sec:problem}

We will study the following three problems in this note.
\begin{enumerate}
\item {\em Neural Networks Architecture Evolution.} We provide a survey
on the architecture evolution of neural networks, including
computational neurons, multi-layer perceptrons (MLPs) and CNNs.
\item {\em Signal Analysis via Multi-Layer RECOS Transform.} We point
out the differences between the single- and multi-layer signal analysis
approaches and explain the working principle of the multi-layer RECOS
transform. 
\item {\em Network Initialization and Guided Anchor Vector Update.} We
examine the CNN initialization scheme carefully since it can be viewed
as an unsupervised clustering and the CNN self-organization property can
be explained.  The supervised learning is achieved by BP using data
labels in the training stage. It will be interpreted as guided anchor
vector update. 
\end{enumerate}

\section*{\bf SOLUTION}\label{sec:solution}

\subsection{CNN Architecture Evolution}\label{sec:A}

We divide the architectural evolution of CNNs into three stages and
provide a brief survey below. 

{\bf Computational Neuron.} A computational neuron (or simply neuron) is
the basic operational unit in neural networks.  It was first proposed by
McCulloch and Pitts in \cite{mcculloch1943logical} to model the
``all-or-none" character of nervous activities. It conducts two
operations in cascade: an affine transform of input vector ${\bf x}$
followed by a nonlinear activation function.  Mathematically, we can
express it as
\begin{equation}\label{eq:neuron}
y=f(b), \quad b=T_{\bf a} ({\bf x})=\sum_{n=1}^N a_n x_n + a_0 \mu = 
{\bf a}^T {\bf x} + a_0 \mu = {\bf a'}^T {\bf x'},
\end{equation}
where $y$, ${\bf x}=(x_1, \cdots, x_N)^T \in R^N$ and ${\bf a}= (a_1,
\cdots, a_N)^T \in R^N$ are the scalar output, the $N$-dimensional input
and model parameter vectors, respectively, $a_0 \mu$ is a bias term with
$\mu=\frac{1}{N}\sum_{n=1}^N x_n$ ({\em i.e.} the mean of all input
elements), $f(.)$ denotes a nonlinear activation function and $b$ is the
intermediate result between the two operations. 

The function, $f(.)$, was chosen to be a delayed step function in form
of $f(b)=u(b-\phi)$ in \cite{mcculloch1943logical}, where $f(b)=1$ if $b
\ge \phi$ and $0$ if $b < \phi$. A neuron is on (in the one-state) if
the stimulus, $b$, is larger than threshold $\phi$.  Otherwise, it is
off (in the zero-state). Multiple neurons can be flexibly connected into
logic networks as models in theoretical neurophysiology. 

For the vision problem, input ${\bf x}$ denotes an image (or image
patch). The neuron should not generate a response for a flat patch since
it does not carry any visual pattern information.  Thus, we set $b=0$ if
all of its elements are equal to a non-zero constant.  It is then
straightforward to derive $\sum_{n=1}^N a_n + a_0 = 0$ or
$a_0=-\sum_{n=1}^N a_n$ is a dependent variable.  We can form augmented
vectors ${\bf x'}=(\mu, x_1, \cdots, x_N)^T \in R^{N+1}$ and ${\bf
a'}=(a_0,a_1,\cdots, a_N)^T \in R^{N+1}$ for ${\bf x}$ and ${\bf a}$,
respectively.  Without loss of generality, we assume $\mu=0$ in the
following discussion. If $\mu \ne 0$, we can either consider the
augmented vector space of ${\bf x'}$ or normalize input ${\bf x}$ to be
a zero-mean vector before the processing and add the mean back after the
processing. 

\begin{figure}
\centering
\includegraphics[width=0.4\linewidth]{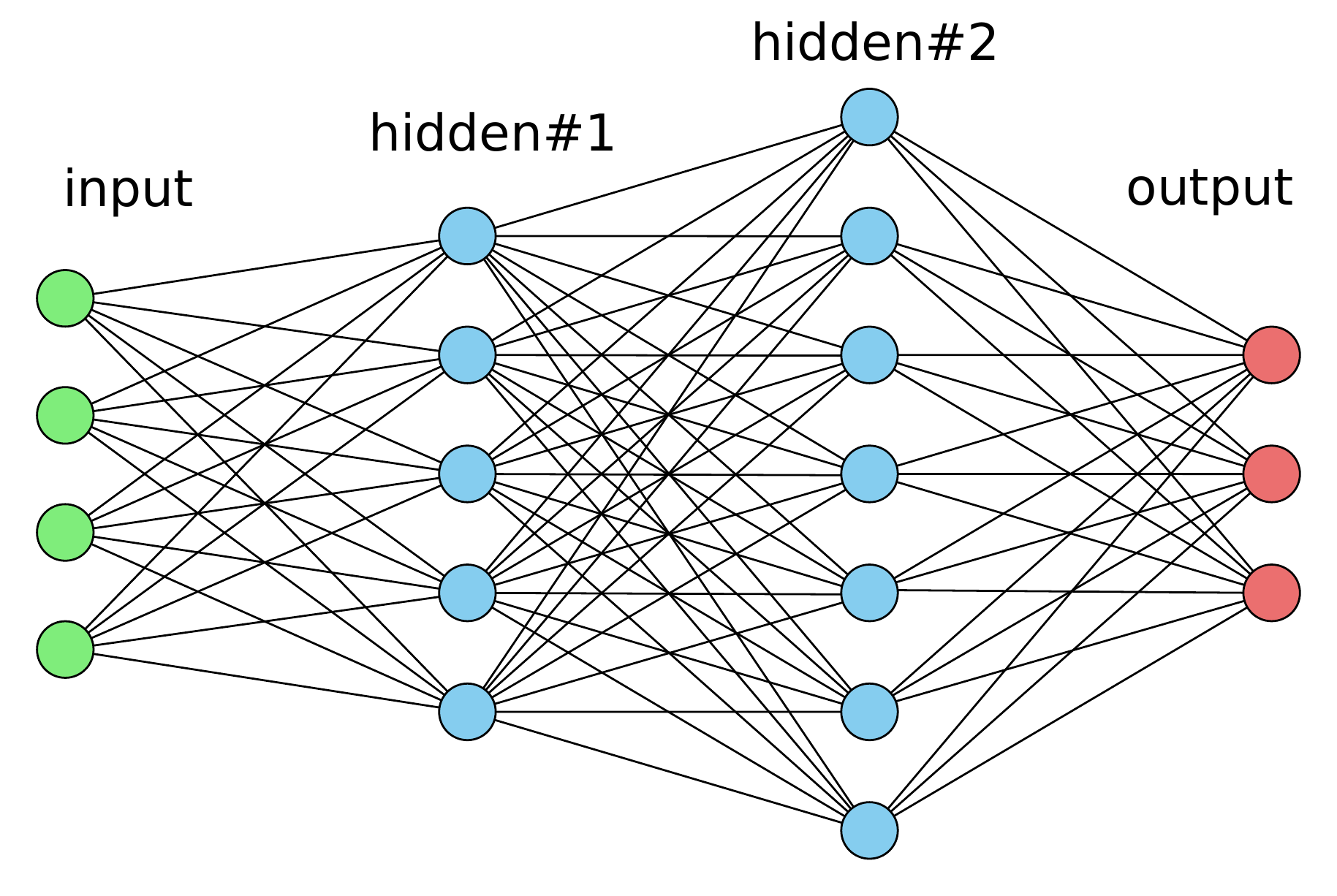}
\caption{An exemplary MLP with one input layer, two hidden layers 
and one output layer.}\label{fig:2GNN}
\end{figure}

{\bf Multi-Layer Perceptrons (MLPs).} The perceptron was introduced by
Rosenblatt in \cite{rosenblatt1957}. One can stack multiple perceptrons
side-by-side to form a perceptron layer, and cascade multiple perceptron
layers into one network. It is called the multi-layer perceptron (MLP)
or the feedforward neural network (FNN).  An exemplary MLP is shown in
Fig.  \ref{fig:2GNN}.  In general, it consists of a layer of input nodes
(the input layer), several layers of intermediate nodes (the hidden
layers) and a layer of output nodes (the output layer). These layers are
indexed from $l=0$ to $L$, where the input and output layers are indexed
with $0$ and $L$, and the hidden layers are indexed from $l=1 \cdots
L-1$, respectively.  Suppose that there are $N_l$ nodes at the $l$th
layer.  Each node at the $l$th layer takes all nodes in the $(i-1)$th
layer as its input.  For this reason, it is called the fully connected
layer.  Clearly, the MLP is end-to-end fully connected.  A modern CNN
often contains an MLP as its building module. 

MLPs were studied intensively in 80s and 90s as decision networks for
pattern recognition applications. The input and output nodes represent
selected features and classification types, respectively.  There are two
major advances from simple neuron-based logic networks to MLPs.  First,
there was no training mechanism in the former since they were not
designed for the machine learning purpose. The BP technique was
introduced in MLPs as a training mechanism for supervised learning.
Since differentiation is needed in the BP yet the step function is not
differentiable, other nonlinear activation functions are adopted in
MLPs.  Examples include the sigmoid function, the rectified linear unit
(ReLU) and the parameterized ReLU (PReLU).  Second, MLPs have a
modularized structure (i.e. perceptron layers) that is suitable for
parallel processing.

As compared with traditional pattern recognition techniques based on
simple linear analysis ({\em e.g}, linear discriminant analysis,
principal component analysis, etc.), MLPs provide a more flexible
mapping from the feature space to the decision space, where the
distribution of feature points of one class can be non-convex and
irregular. It is built upon a solid theoretical foundation proved by
Cybenko \cite{cybenko1989approximation} and Hornik {\em et al.}
\cite{hornik1989multilayer}. That is, a network with only one hidden
layer can be a universal approximator if there are ``enough" neurons. 

\begin{figure}
\centering
\includegraphics[width=0.7\linewidth]{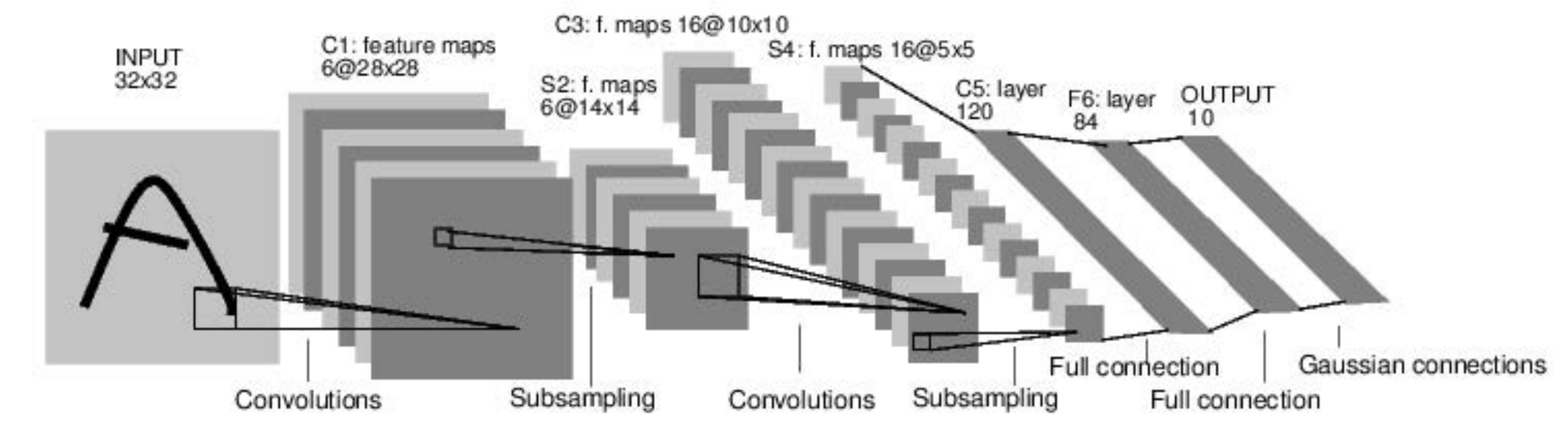}
\caption{The LeNet-5 architecture \cite{LeNet1998} as an exemplary CNN.}\label{fig:LeNet}
\end{figure}

{\bf Convolutional Neural Networks (CNNs).} Fukushima's Neocognitron
\cite{fukushima1980} can be viewed as an early form of a CNN.  The
architecture introduced by LeCun {\em et al.} in \cite{LeNet1998} serves
as the basis of modern CNNs.  The main difference between MLPs and CNNs
lies in their input space - the former are features while the latter are
source data such as image, video, speech, etc. This is not a trivial
difference.  Let us use the LeNet-5 shown in Fig.  \ref{fig:LeNet} as an
example, whose input is an image of size 32 by 32.  Each pixel is an
input node.  It would be very challenging for an MLP to handle this
input since the dimension of the input vector is $32\times 32=1,024$.
The diversity of possible visual patterns is huge.  As explained later,
the nodes in the first hidden layer should provide a good representation
for the input signal.  Thus, it implies a large number of nodes in
hidden layers. The number of links (or filter weights) between the input
and the first hidden layers is $N_0 \times N_1$ due to full connection.
This number can easily go to the order of millions. If the image
dimension is in the order of millions such as those captured by the
smart phones nowadays, the solution is clearly unrealistic. 

Instead of considering interactions of all pixels in one step as done in
the MLP, the CNN decomposes an input image into smaller patches, known
as receptive fields for nodes at certain layers. It gradually enlarges
the receptive field to cover a larger portion of the image. For example,
the filter size of the first two convolutional layers of LeNet-5 is
$5\times5$.  The first convolutional layer considers interactions of
pixels in the short range. Since the patch size is small, the diversity
is less. One can use 6 filters to provide a good approximation to the
$5\times5$ source patches, and all source patches share the same 6
filters regardless of their spatial location.  After subsampling, the
second convolutional layer examines interaction of pixels in the
mid-range. After another subsampling, the whole spatial domain is
shrinked to a size $5\times5$ so that it can take global interaction
into account using full connection. Typically, the interaction contains
not only spatial but also spectral elements (e.g.  the RGB three
channels and multiple filter responses at the same spatial location) and
all interactions are modeled by computational neurons as given in Eq.
(\ref{eq:neuron}). 

It is typical to decompose a CNN into two sub-networks: the feature
extraction (FE) subnet and the decision making (DM) subnet. The FE
subnet consists of multiple convolutional layers while the DM subnet is
composed by a couple of fully connected layers. Roughly speaking, the FE
subnet conducts clustering aiming at a new representation through a
sequence of RECOS transforms. The DM subnet links data representations
to decision labels, which is similar to the classification role of MLPs.
The exact boundary between the FE subnet and the DM subnet is actually
blurred in the LeNet-5. It can be either S4 or C5 . If we view S4 as the
boundary, then C5 and F6 are two hidden layers of the DM subnet. On the
other hand, if we choose C5 as the boundary, then there is only one
hidden layer ({\em i.e.,} F6) in the DM subnet.  Actually, since these
two subnets are connected side-by-side, the transition from the
representation to the classification happens gradually and smoothly. 

One main advantage of CNNs over the state vector machine (SVM) and the
random forest (RF) classifiers is that the feature extraction task is
automatically done through the BP from the last layer to the first
layer.  Generally speaking, discriminant features are difficult to find
for traditional classifiers such as the MLP, SVM and RF.  Such tasks,
called the feature engineering, demand the domain knowledge.
Furthermore, it is difficult to argue that ad hoc features found
empirically are optimal in any sense. This explains why the traditional
computer vision field is fragmented by different applications. After the
emergence of CNNs, the domain knowledge is no more important in feature
extraction, yet it plays a critical role in data labeling (known as the
label engineering).  To give an example, anaconda, vipers, titanoboa,
cobras, rattlesnake, etc. are finer classifications of snakes. It
demands expert knowledge to collect and label their images.  The CNN
provides a powerful tool in data-driven supervised learning, where the
emphasis is shifted from ``extracting features from the source data" to
``constructing datasets by pairing carefully selected data and their
labels". 

\subsection{Single-Layer RECOS Transform}\label{sec:B}

Our discussion applies to ${\bf x}$ and ${\bf a}$ if $\mu=0$ or
augmented vectors ${\bf x'}$ and ${\bf a'}$ if $\mu \neq 0$.  For
convenience, we only consider the case with $\mu=0$. The generalization
to $\mu \ne 0$ is straightforward. 

{\bf Clustering on Sphere's Surface.} There is a modern interpretation
to the function of a single perceptron layer based on the clustering
notion.  Since data clustering is a well understood discipline, one can
understand the operation of CNNs better if a connection between the
operation of a perceptron layer and data clustering can be established.
This link was built in \cite{kuo2016understanding}. It will be
repeated below. Let
$$
S=\left\{ {\bf x} \biggr\rvert ||{\bf x}||=1 \right\}.
$$
be an $N$-dimensional unit hyper-sphere (or simply sphere). We consider
clustering of points in $S$ using the geodesic distance.  For an
arbitrary vector ${\bf x}$ to be a member in $S$, we need to normalize
it by its magnitude $g=||{\bf x}||$.  If ${\bf x}$ is an image patch,
the magnitude normalization after its mean removal has a physical
meaning; namely, contrast adjustment.  When $g$ is smaller than a
threshold, the patch is nearly flat. A flat patch carries little visual
information yet its normalization does amplify noise. In this case, it
is better to treat it as a zero vector. When $g$ is larger than the
threshold, vector ${\bf x}$ does represent a visual pattern and humans
perceive little difference between the original and normalized patches
since the contrast has little effect on visual patterns.  Although this
normalization procedure is not implemented in today's CNNs, the
following mathematical analysis can be significantly simplified while
the essence of CNNs can still be well captured. 

The geodesic distance of two points, ${\bf x}_i$ and ${\bf x}_j$ in $S$,
is proportional to the magnitude of their angle, which can be computed by
$$
\theta ({\bf x}_i, {\bf x}_j) = \cos^{-1} ({\bf x}^T_i {\bf x}_j).
$$
Since $\cos \theta$ is a monotonically decreasing function for $0^o \le
|\theta| \le 180^o$, we can use the correlation, $0 \le {\bf x}_i^T {\bf
x}_j = \cos \theta \le 1$, as another distance measure between them,
and cluster vectors in $S$ accordingly.  Note that, when $90^o \le
|\theta| \le 180^o$, the correlation, ${\bf x}^T_i {\bf x}_j = \cos
\theta$, is a negative value. 

For MLPs and CNNs, a set of neurons is used to operate on a set of input
nodes. For example, nodes in each hidden layer and the output layer in
Fig. \ref{fig:2GNN} take the weighted sum of values of nodes in the
preceding layer as their outputs.  These outputs are treated as one
inseparable unit which becomes the input to the next layer.  Each neuron
has a filter weight vector denoted by ${\bf a}_k$, $k=1, \cdots, K$.  In
the signal processing terminology, the set of neurons forms a filter
bank.  The ``REctified-COrrelations on a Sphere" (RECOS) model
\cite{kuo2016understanding} describes the relationship between nodes of
the $(l-1)$th and $l$th layers, $l=1, 2, \cdots$, where the input layer
is the $0$th layer. There are three RECOS units in cascade for the MLP
in Fig.  \ref{fig:2GNN}. One corresponds to a filter bank.  The filter
weight vector is called an anchor vector since it serves as a reference
pattern associated with a neuron unit. 

{\bf Need of Rectification.} A neuron computes the correlation between
an input vector and its anchor vector to measure their similarity.
There are $K$ neurons in one RECOS unit. The projection of ${\bf x}$
onto all anchor vectors, ${\bf a}_k$, can be written in form of
$$
{\bf y} = {\bf A} {\bf x}, \quad {\bf A}^T = [{\bf a}_1 \cdots {\bf a}_k 
\cdots {\bf a}_K],
$$
where ${\bf y}=(y_1, \cdots, y_k, \cdots y_K)^T \in R^K$, $y_k={\bf
a}_k^T{\bf x}$ and ${\bf A} \in R^{K\times N}$.  For input vectors ${\bf
x}_i$ and ${\bf x}_j$, their corresponding outputs are ${\bf y}_i$ and
${\bf y}_j$. If the geodesic distance of ${\bf x}_i$ and ${\bf x}_j$ in
$S$ is close, we expect the distance of ${\bf y}_i$ and ${\bf y}_j$ in
the $K$-dimensional output space to be close as well. 

\begin{figure}
\centering
\includegraphics[width=0.25\linewidth]{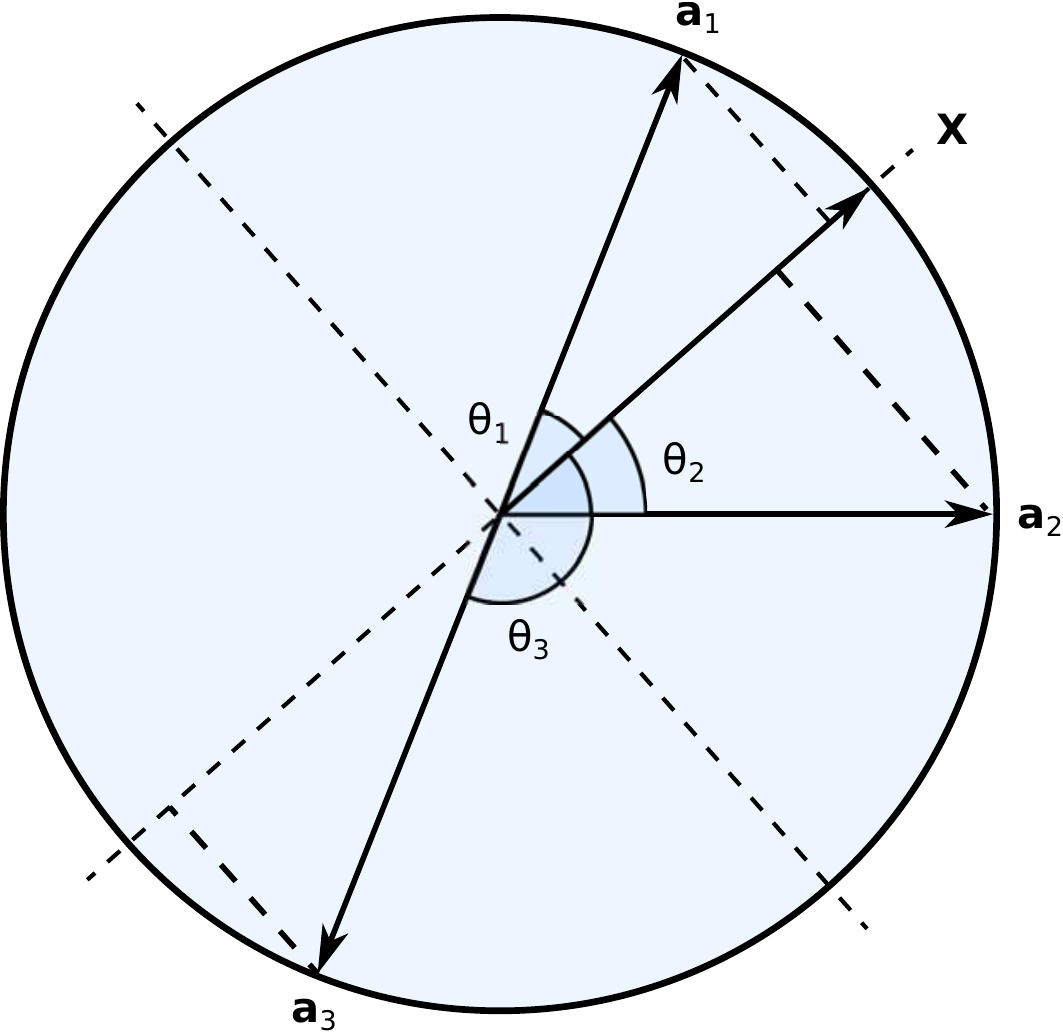}
\caption{Illustrate the need of rectification \cite{kuo2016understanding}.}
\label{fig:rectification}
\end{figure}

To show the necessity of rectification, a 2D example is illustrated in
Fig.  \ref{fig:rectification}, where ${\bf x}$ and ${\bf a}_k$
($k=1,2,3$) denote an input and three anchor vectors on the unit circle,
respectively, and $\theta_i$ is their respective angle.  Since
$\theta_1$ and $\theta_2$ are less than 90 degrees, ${\bf a}^T_1 {\bf
x}$ and ${\bf a}^T_2 {\bf x}$ are positive.  The angle, $\theta_3$, is
larger than 90 degrees and correlation ${\bf a}^T_3 {\bf x} $ is
negative. The two vectors, ${\bf x}$ and ${\bf a}_3$, are far apart in
terms of the geodesic distance. Since $\cos \theta$ is monotonically
decreasing for $0^o \le |\theta| \le 180^o$, it can be used to reflect
the order of the geodesic distance in one layer. 

However, when two RECOS units are in cascade, the filter weights of the
2nd RECOS unit can take positive or negative values. If the response of
the 1st RECOS unit is negative, the product of a negative response and a
negative filter weight will produce a positive value. On the other hand,
the product of a positive response and a positive filter weight will
also produce a positive value. If the nonlinear activation unit did not
exist, the cascaded system would not be able to differentiate them. For
example, the geodesic distance of ${\bf x}$ and ${\bf -x}$ should be
farthest. However, they yield the same result and their original patches
become indistinguishable under this scenario.  Similarly, a system
without rectification cannot differentiate the following two cases: 1) a
positive response at the first layer followed by a negative filter
weight at the second layer; and 2) a negative response at the first
layer followed by a positive filter weight at the second layer. 

\begin{figure}
\centering
\begin{subfigure}[b]{0.2\linewidth}
\centering
\includegraphics[width=\linewidth]{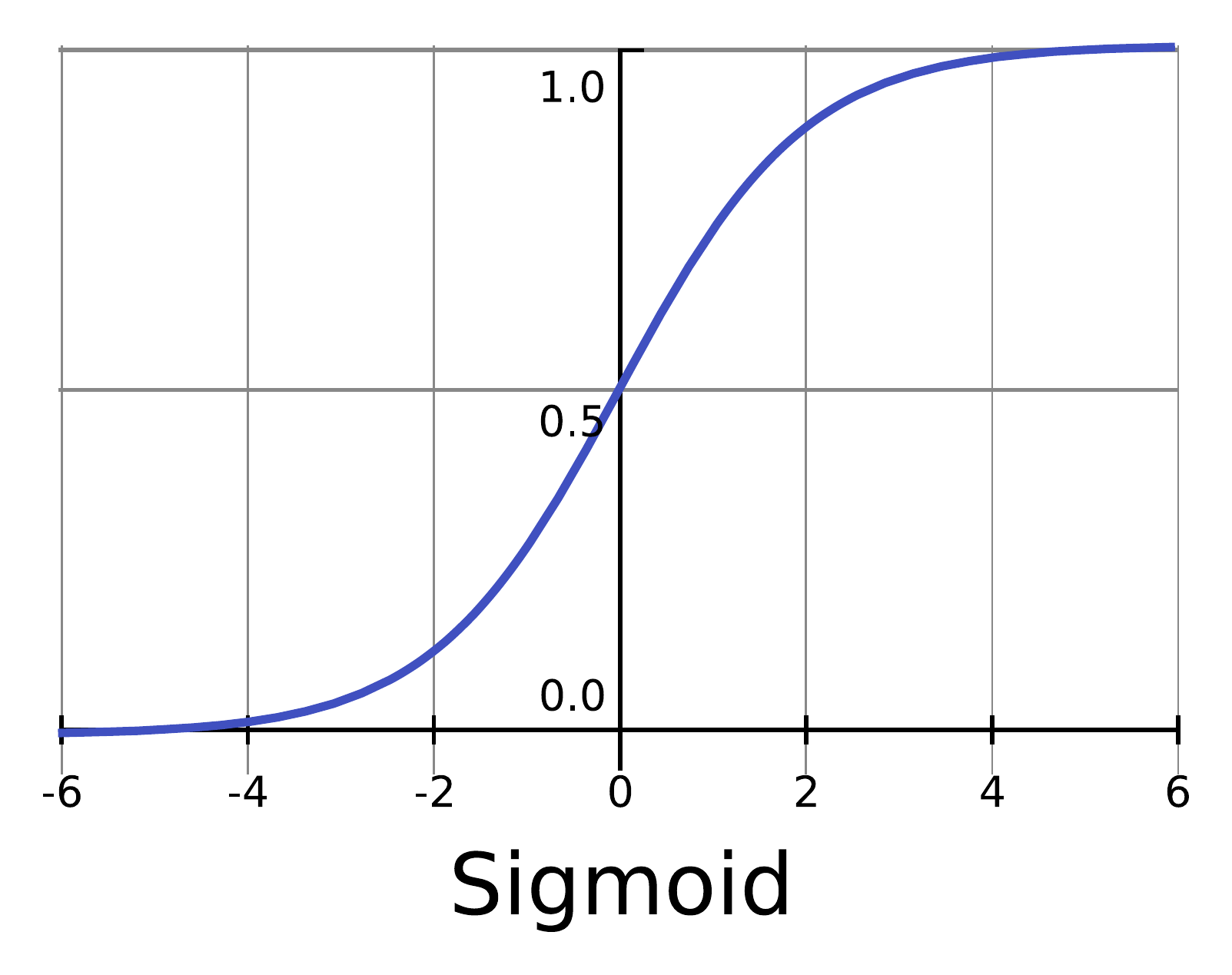}
\caption{}
\end{subfigure}
\begin{subfigure}[b]{0.2\linewidth}
\centering
\includegraphics[width=\linewidth]{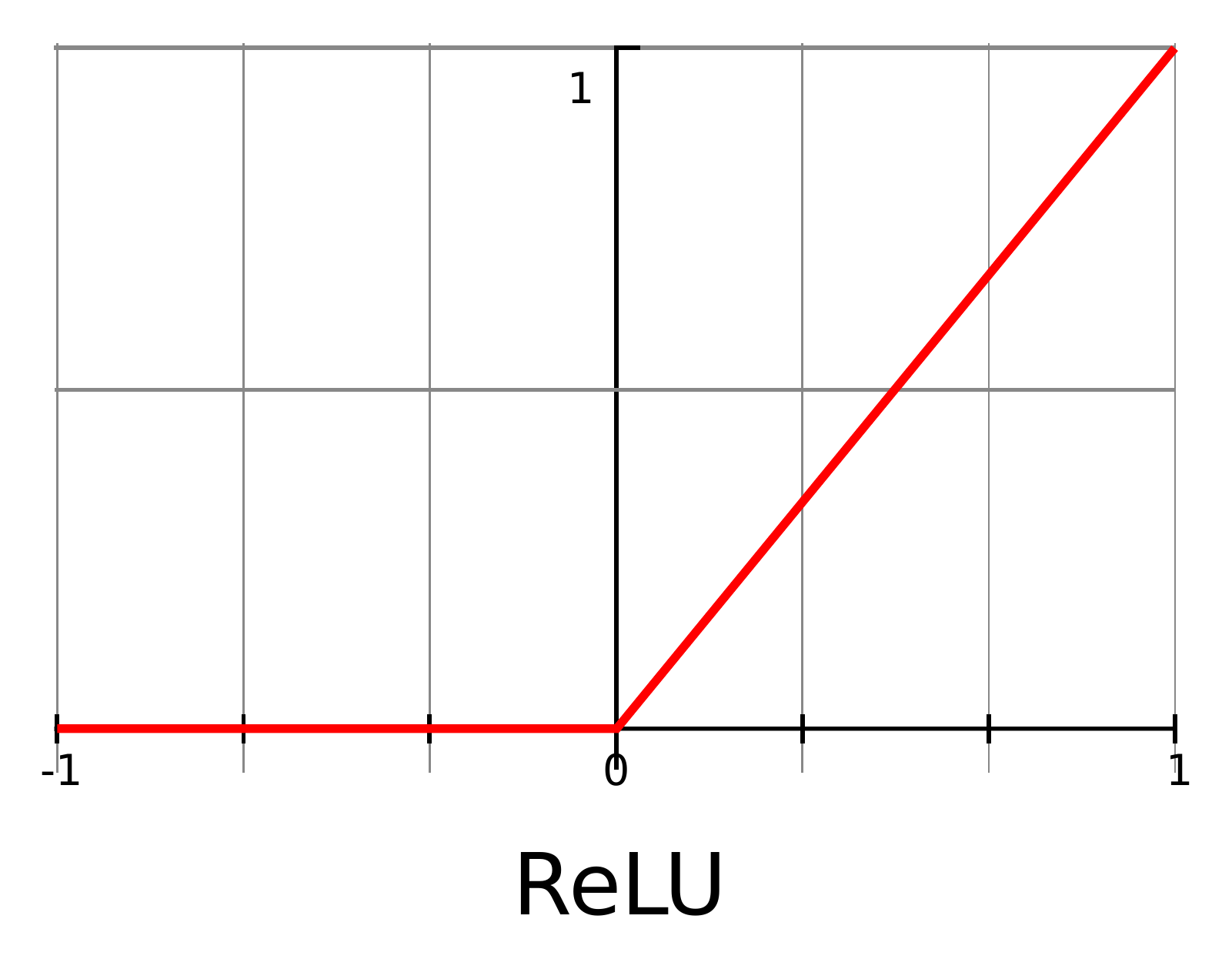}
\caption{}\end{subfigure}
\begin{subfigure}[b]{0.2\linewidth}
\centering
\includegraphics[width=\linewidth]{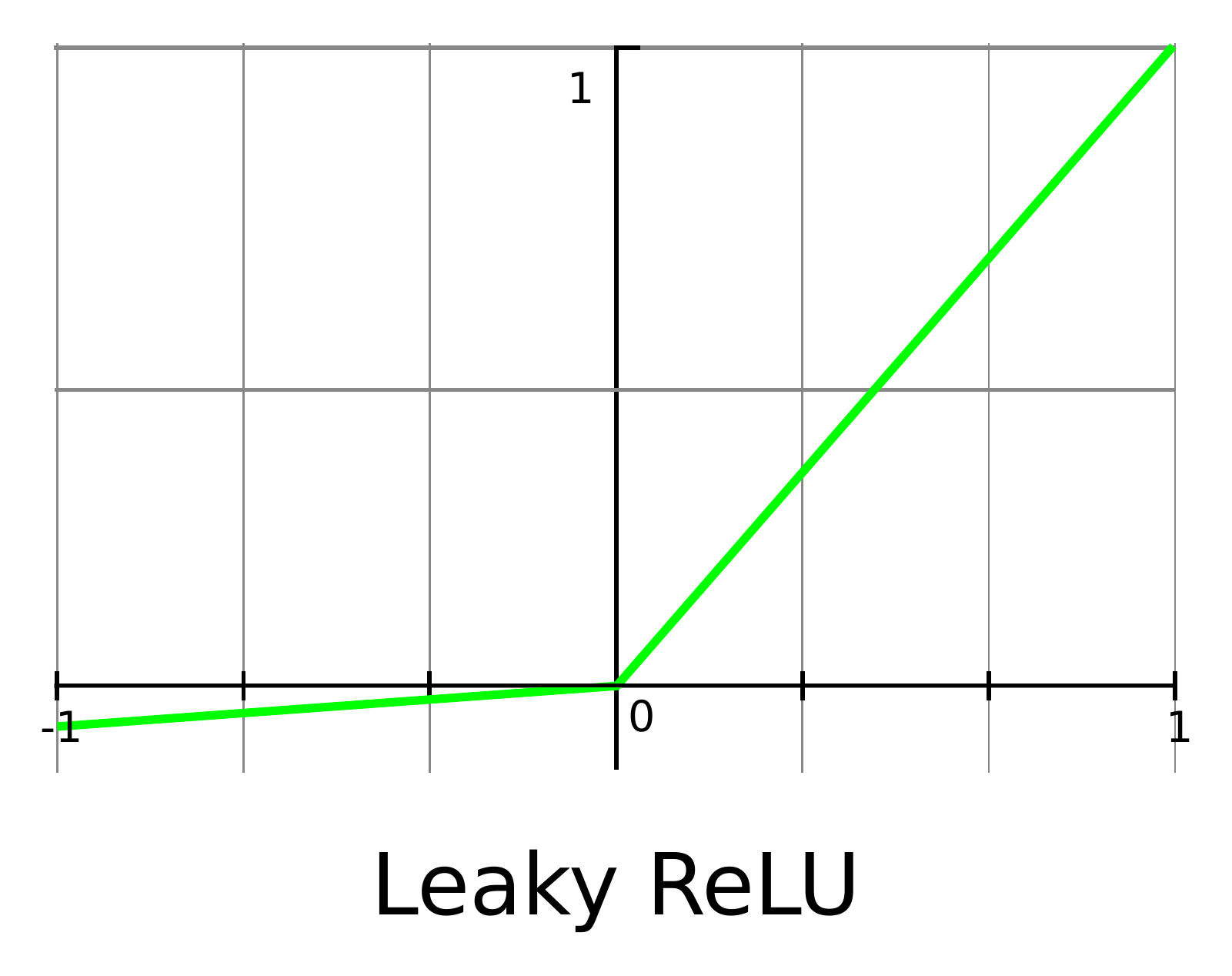}
\caption{}\end{subfigure}
\begin{subfigure}[b]{0.2\linewidth}
\centering
\includegraphics[width=\linewidth]{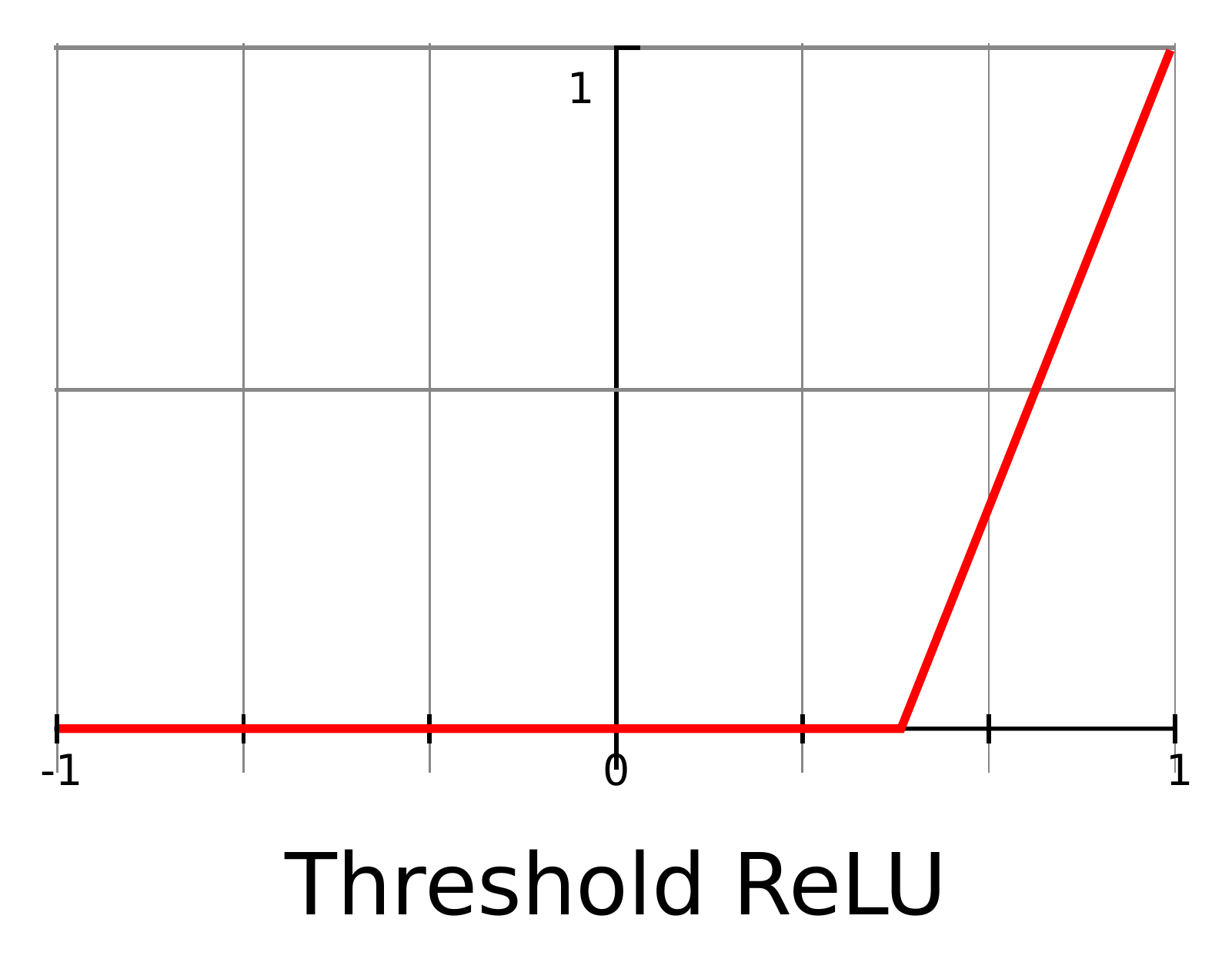}
\caption{}\end{subfigure}
\caption{Illustration of four rectifiers: (a) the sigmoid function,
(b) the ReLU (middle), (c) the Leaky ReLU and (d) the Threshold ReLU.}
\label{fig:clipping}
\end{figure}

{\bf Rectifier Design.} Since a nonlinear activation unit is used to
rectify correlations, it is called a rectifier here.  To avoid the
above-mentioned confusion cases, we impose the following two
requirements on a rectifier. 
\begin{enumerate}
\item The output ${\bf a}^T_k {\bf x}$ should be rectified to be a
non-negative value. 
\item The rectification function should be monotonically increasing so
as to preserve the order of the geodesic distance. 
\end{enumerate}
Three rectifiers are often used in MLPs and CNNs.  They are the sigmoid
function, the rectified linear unit (ReLU) and the parameterized ReLU
(PReLU) as shown in Figs.  \ref{fig:clipping}(a)-(c).  The PReLU is also
known as the leaky ReLU.  Both the sigmoid and ReLU satisfy the above
two requirements.  Although the PReLU does not satisfy the first
requirement strictly, it does not have a severe negative impact on
spherical surface clustering.  This is because a negative correlation is
rectified to a significantly smaller negative value. 

Based on the two requirements, one can design different rectifiers.  One
example is shown in Fig.  \ref{fig:clipping}(d). It is called the
threshold ReLU (TReLU). The rectification function can be defined as
$\mbox{TReLU}(x)=0$, if $x < \phi$ and
$\mbox{TReLU}(x)=\frac{x-\phi}{1-\phi}$ if $x \geq \phi$. When $\phi=0$,
TReLU is reduced to ReLU. For the LeNet-5 applied to the MNIST dataset,
we observe better performance as $\phi$ increases from 0 to 0.5 and then
decreases. One advantage of $\mbox{TReLU}(\phi)$ with $\phi>0$ is that
we can block the influence of more anchor vectors. When $\phi=0$, we
block the influence of anchor vectors that have an angle larger than 90
degrees with respect to the input vector. When $\phi=0.5$, we block the
influence of anchor vectors that have an angle larger than 60 degrees.
The design of an optimal rectifier for target applications remains to be
an open problem. 

\subsection{Multi-layer RECOS Transform}\label{sec:C}

{\bf Single-Layer Signal Analysis via Representation.} Signal modeling and
representation is commonly used in the signal processing field for
signal analysis. Typically, we have a linear model in form of
\begin{equation}\label{eqn:model-A}
{\bf x}={\bf A} {\bf c},
\end{equation}
where ${\bf x}\in R^N$ denotes the signal of interest, ${\bf A} \in
R^{N\times M}$ is a representation matrix and ${\bf c}\in R^M$ is the
coefficient vector. If $M=N$ and the column vectors of ${\bf A}$ form a
set of basis functions, Eq. (\ref{eqn:model-A}) defines a transform from
one basis to another.  The task is in selecting powerful basis functions
to represent signals of interest. Fourier and wavelet transforms are
well known examples. Then, a subset of coefficient vector ${\bf c}$ can
be used as the feature vector.  If $M > N$, there exist infinitely many
solutions in ${\bf c}$.  We can impose constraints on ${\bf c}$, leading
to the linear least-squares solution, sparse coding, among others. For
the sparse representation, the task is in finding a good dictionary,
${\bf A}$, to represent the underlying signal effectively.  Again, a
subset of coefficient vector ${\bf c}$ can be chosen as features. 

{\bf Multi-Layer Signal Analysis via Cascaded Transforms.} The CNN
approach provides a brand new framework for signal analysis. Instead of
finding a representation for signal analysis, it relies on a sequence of
cascaded transforms that builds a link between the input signal space
and the output decision space. The operation at each layer is to conduct
spherical surface's clustering of input samples with a rectified output
(i.e. the RECOS transform). 

For MLPs, each network corresponds to a simple cascade of multiple
RECOS transforms.  Mathematically, we have
\begin{equation}\label{eqn:model-R}
{\bf d}= {\bf B}_L \cdots {\bf B}_{l} \cdots {\bf B}_1 {\bf x},
\end{equation}
where ${\bf x}$ is an input signal, ${\bf d}=(d_1, \cdots, d_c, \cdots,
d_C)$ is an output vector in the decision space indicating the
likelihood in class $c$ with $c=1, \cdots, C$, and ${\bf B}_l$ is the
$l$th layer RECOS transform matrix with $l=1, \cdots, L$.  The input and
output to the $l$th layer RECOS transform ${\bf B}_l$ are denoted by ${\bf
x}_{l-1}$ and ${\bf x}_l$, respectively. Thus, we get
\begin{equation}\label{eqn:model-B}
{\bf x}_l = {\bf B}_l {\bf x}_{l-1}, 
\quad \mbox{where} \quad 
{\bf B}_l = {\bf R} \circ {\bf A}_l,
\end{equation}
and where ${\bf R}$ is the element-wise rectification function operating on
the output of ${\bf A}_l {\bf x}_{l-1}$.  Clearly, we have ${\bf x}_{0}=
{\bf x}$ and ${\bf x}_{L}={\bf d}$. 

The ground truth ${\bf d}$ is that $d_i=1$ if $i$ is the target class
while $d_j=0$ if $j$ is not the target class. It is called the one-hot
vector.  The training samples have both input ${\bf x}$ and its label 
${\bf d}$.  The testing samples have only input ${\bf x}$, and we need
to predict its output ${\bf d}$ and convert it to its nearest one-hot
vector.  The task is in finding good ${\bf B}_l$, $l=1, \cdots, L$, so
as to minimize the classification error. 

For CNNs, we have two types of ${\bf B}_l$ in form of:
\begin{equation}\label{eqn:model-C}
{\bf B}_l^{C} = {\bf P} \bigcup_{s \in \Omega} {\bf R} \circ {\bf A}_{l,s}, 
\quad \mbox{and} \quad
{\bf B}_l^{F} = {\bf R} \circ {\bf A}_l,
\end{equation}
where ${\bf A}_{l,s}$ denotes a convolutional filter at layer $l$ with
spatial index $s$, $\bigcup_s$ is the union of outputs from a
neighborhood, $\Omega$, and ${\bf P}$ denotes a pooling operation.  The
union of outputs from a set of parallel convolutional filters serve as
the input to the filter at the next layer. The two RECOS transforms,
${\bf B}_l^{C}$ and ${\bf B}_l^{F}$, are called the convolutional layer
and the fully connected layer, respectively, in the modern CNN
literature. Clearly, ${\bf B}_l={\bf B}_l^F$ in Eq. (\ref{eqn:model-B}).

It is inspiring to compare the two signal analysis approaches as given
in Eqs. (\ref{eqn:model-A}) and (\ref{eqn:model-R}). The one in Eq.
(\ref{eqn:model-A}) is a single layer approach where no rectification is
needed. The one in Eq. (\ref{eqn:model-R}) is a multi-layer approach and
rectification is essential.  The single layer approach seeks for a
better signal representation. For example, a multiple-scale signal
representation was developed using the wavelet transform.  A sparse
signal representation was proposed using a trained dictionary.  The
objective is to find an ``optimal" representation to separate critical
components in desired signals from others. 

In contrast, the CNN approach does not intend to decompose underlying
signals. Instead, it adopts a sequence of RECOS transforms to cluster
input data based on their similarity layer by layer until the output
layer is reached. The output layer predicts the likelihood of all
possible decisions (e.g., object classes). The training samples provide a
relationship between an image and its decision label. The CNN can
predict results even without any supervision, although the prediction accuracy
would be low.  The training samples guide the CNN to form more suitable
anchor vectors (thus better clusters) and connect clustered data with
decision labels. To summarize, we can express the multi-layer RECOS transform as
\begin{eqnarray*}
\mbox{MLP:} & & {\bf x}={\bf x}_0 \overset{{\bf B}_1^F}{\longrightarrow} {\bf x}_1 
\overset{{\bf B}_2^F}{\longrightarrow} \cdots \overset{{\bf B}_{L-1}^F}{\longrightarrow} 
{\bf x}_{L-1} \overset{{\bf B}_L^F}{\longrightarrow} {\bf x}_L={\bf d}, \\
\mbox{CNN:} & & {\bf x}={\bf x}_0 \overset{{\bf B}_1^C}{\longrightarrow} {\bf x}_1 
\overset{{\bf B}_2^C}{\longrightarrow} \cdots \overset{{\bf B}_{m}^C}{\longrightarrow} 
{\bf x}_{m}  \overset{{\bf B}_{m+1}^F}{\longrightarrow} {\bf x}_{m+1}  
\overset{{\bf B}_{m+2}^F}{\longrightarrow} \cdots \overset{{\bf B}_{L-1}^F}{\longrightarrow} 
{\bf x}_{L-1} \overset{{\bf B}_L^F}{\longrightarrow} {\bf x}_L={\bf d},
\end{eqnarray*}
The output from the $l$th layer, ${\bf x}_l$, serves as the input to the
$(l+1)$th layer. It is called the intermediate representation at the
$l$th layer or the $l$th intermediate representation in short. 

\begin{figure}
\centering
\includegraphics[width=0.4\linewidth]{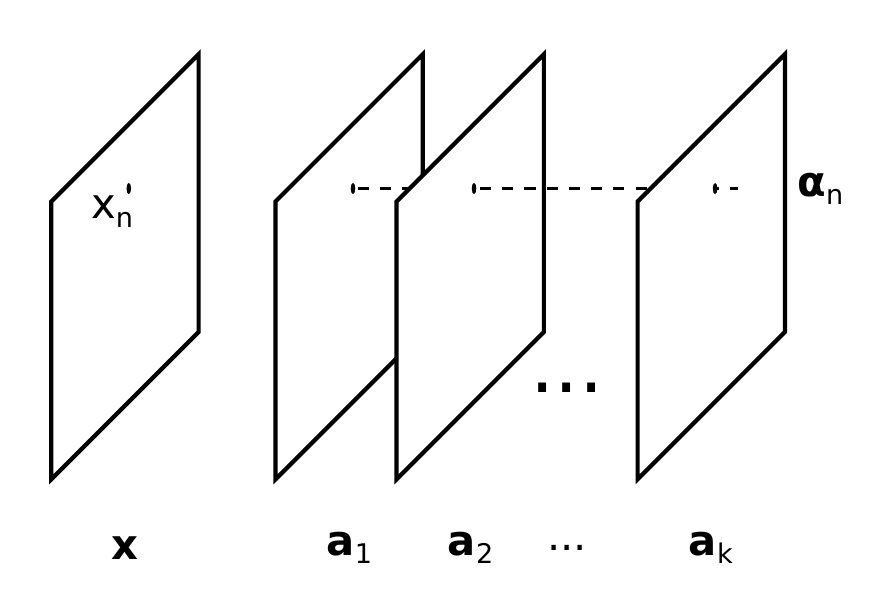}
\caption{Visualization of anchor-position vector ${\boldsymbol \alpha}_n$ 
\cite{kuo2016understanding}.}\label{fig:alpha}
\end{figure}

It is important to have a deeper understanding on the compound effect of
two RECOS transforms in cascade. This was thoroughly studied in
\cite{kuo2016understanding}, and the main result is summarized below. A
representative 2D input and its corresponding anchor vectors are shown
in Fig.  \ref{fig:alpha}.  Let ${\boldsymbol \alpha}_n$ be a
$K$-dimensional vector formed by the same position (or element) of ${\bf
a}_k$. It is called the anchor-position vector since it captures the
position information of anchor vectors.  Although anchor vectors ${\bf
a}_k$ capture global representative patterns of ${\bf x}$, they are weak
in capturing position sensitive information. This shortcoming can be
compensated by modulating outputs with elements of anchor-position
vector ${\boldsymbol \alpha}_n$ in the next layer. 

Let us use layers S4, C5 and F6 in LeNet-5 as an example.  There are
$120$ anchor vectors of dimension $400$ from S4 to C5. We collect $400$
anchor-position vectors of dimension $120$, multiply the output at C5 by
them to form a set of modulated outputs, and then compute $84$ anchor
vectors of dimension $120$ from C5 to F6.  Note that the output at C5
contains primarily the spectral information but not the position
information. If a position in the input vectors has less consistent
information, the variance of its associated anchor position vector will
be larger and the modulated output will be more random. As a result, its
impact on the formation of the $84$ anchor vectors is reduced.  For more
details, we refer to the discussion in \cite{kuo2016understanding}. 

{\bf New Clustering Representation.} We have a one-to-one association
between a data sample and its cluster in traditional clustering schemes.
However, this is not the case in the RECOS transform. A new clustering
representation is adopted by MLPs and CNNs. That is, for an input vector ${\bf
x}$, the RECOS transform generates a set of $K$ non-negative correlation
values as the output vector of dimension $K$.  This representation
enables repetitive clustering layer by layer as given in Eq.
(\ref{eqn:model-B}). For an input, one can determine the significance of
clusters according to the magnitude of the rectified output value. If
its magnitude for a cluster is zero, ${\bf x}$ is not associated with
that cluster. A cluster is called a relevant or irrelevant one depending
on whether it has an association with ${\bf x}$. Among all relevant ones, 
we call cluster $i$ the ``primary" cluster for input ${\bf x}$ if
$$
i=\mbox{arg} \max_k {\bf a}_k^T {\bf x}.
$$
The remaining relevant ones are ``auxiliary" clusters.

The FE subnet uses anchor vectors to capture local, mid-range and
long-range spatial patterns. It is difficult to predict the clustering
structure since new information is introduced at a new layer.  The DM
subnet attempts to reduce the dimension of intermediate representations
until it reaches the dimension of the decision space. We observe that
the clustering structure becomes more obvious as the layer of the DM
subnet goes deeper. That is, the output value from the primary cluster
is closer to unity while the number of auxiliary clusters is fewer and
their output values become smaller. When this happens, an anchor vector
provides a good approximation to the centroid for the corresponding
cluster. 

The choice of anchor vector numbers, $K_l$, at the $l$th layer is an
important problem in the network design. If input data ${\bf x}_{l-1}$
has a clear clustering structure (say, with $h$ clusters), we can set
$K_l=h$.  However, this is often not the case. If $K_l$ is set to a
value too small, we are not able to capture the clustering structure of
${\bf x}_{l-1}$ well, and it will demand more layers to split them.  If
$K_l$ is set to a value too large, there are more anchor vectors than
needed and a stronger overlap between rectified output vectors will be
observed. As a result, we still need more layers to separate them.
Another way to control the clustering process is the choice of the
threshold value, $\phi$, of the TReLU. A higher threshold value can
reduce the negative impact of a larger $K_l$ value. The tradeoff between
$\phi$ and $K_l$ is an interesting future research topic. 

The interpretation of a CNN as a guided multi-layer RECOS transform is
helpful in understanding the CNN self-organization capability as
discussed below. 

\subsection{Network Initialization and Guided Anchor Vector Update}\label{sec:D}

Data clustering plays a critical role in the understanding of the
underlying structure of data. The k-means algorithm, which is probably
the most well-known clustering method, has been widely used in pattern
recognition and supervised/unsupervised learning.  As discussed earlier,
each CNN layer conducts data clustering on the surface of a
high-dimensional sphere based on a rectified geodesic distance.  Here,
we would like to understand the effect of multiple layers in cascade
from the input data source to the output decision label.  For
unsupervised learning such as image segmentation, several challenges
exist in data clustering \cite{jain2010data}.  Questions such as ``what
is a cluster?" ``how many clusters are present in the data?" ``are the
discovered clusters and partition valid?" remain open.  These questions
expose the limit of unsupervised data clustering methods. 

In the context of supervised learning, traditional feature-based methods
extract features from data, conduct clustering in the feature space and,
finally, build a connection between clusters and decision labels.
Although it is relatively easy to build a connection between the data
and labels through features, it is challenging to find effective
features.  In this setting, the dimension of the feature space is
usually significantly smaller than that of the data space. As a
consequence, it is unavoidable to sacrifice rich diversity of input
data. Furthermore, the feature selection process is guided by humans
based on their domain knowledge (i.e. the most discriminant properties
of different objects).  This process is heuristic. It can get overfit
easily.  Human efforts are needed in both data labeling and feature
design. 

CNNs offer an effective supervised learning solution, where supervision
is conducted by a training process using data labels. This supervision
closes the semantic gap between low-level representations ({\em e.g.},
the pixel representation) and high-level semantics.  Furthermore, the
CNN self-organization capability was well discussed in 80s and 90s,
e.g., \cite{fukushima1980}.  By self-organization, the network can learn
with little supervision.  To put the above two together, we expect that
CNNs can provide a wide range of learning paradigms -- from the
unsupervised, weakly supervised to heavily supervised learning.  This
intuition can be justified by considering a proper anchor vector
initialization scheme (for self-organization) and providing proper
``guidance" to the proposed multi-layer RECOS transform in anchor vector
update (for supervised learning). 

{\bf Network Initialization.} The CNN conducts a sequence of
representation transforms using cascaded RECOS units. The dimension of
transformed representations gradually decreases until it reaches the
number of output classes. Since labels of output classes are provided by
humans with a semantic meaning, the whole end-to-end process is called
the {\em guided} (or supervised) transform. 

Before examining the effect of label guidance, we first compare two
network initialization schemes: 1) the random initialization and 2) the
k-means initialization. For the latter, we perform k-means at each layer
based on its corresponding input data samples (with zero-mean and
unit-length normalization), and repeat this process from the input to
the output layer after layer.  Random initialization is commonly adopted
nowadays. Based on the above discussion, we expect the k-means
initialization to be a better choice.  This is verified by our
experiments in the LeNet-5 applied to the MNIST dataset. 

\begin{figure}
\centering
\includegraphics[width=0.4\linewidth]{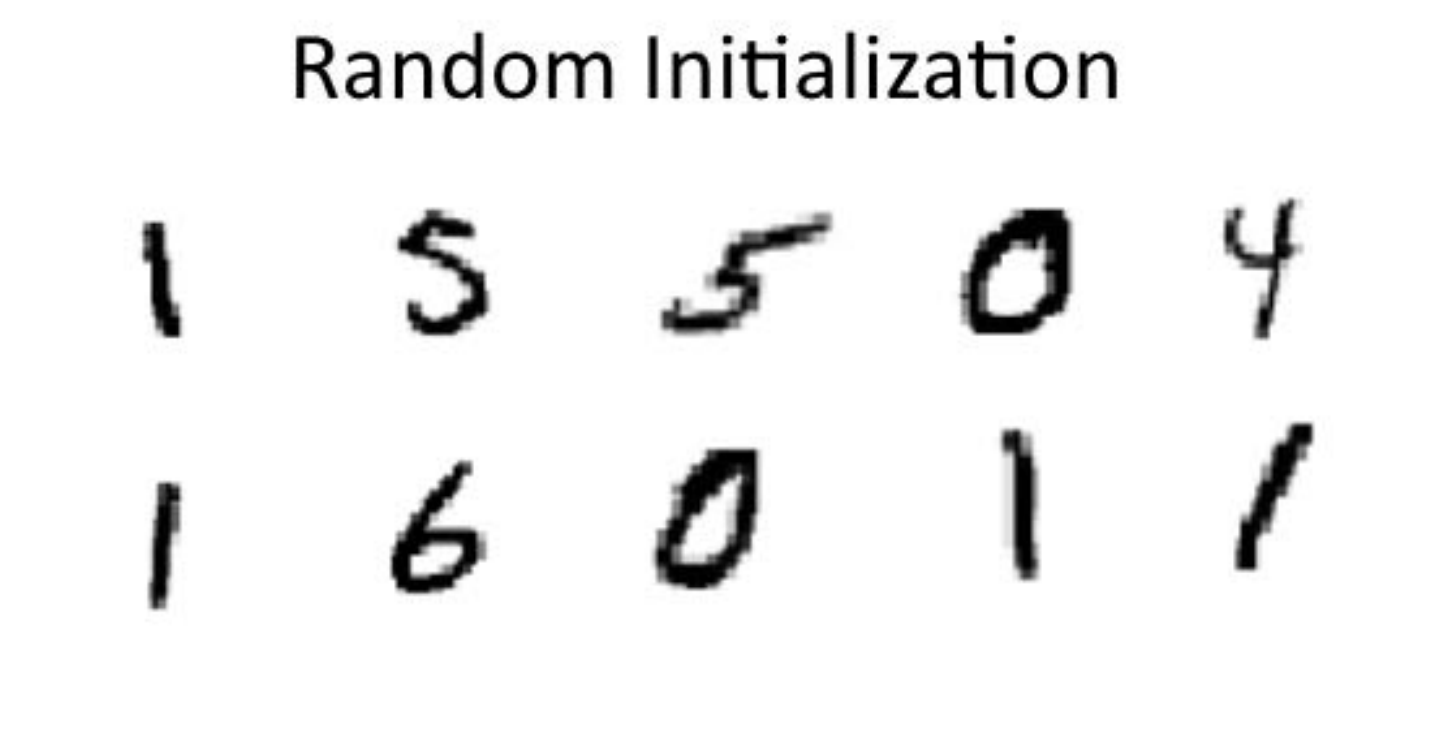} \hspace{1cm}
\includegraphics[width=0.4\linewidth]{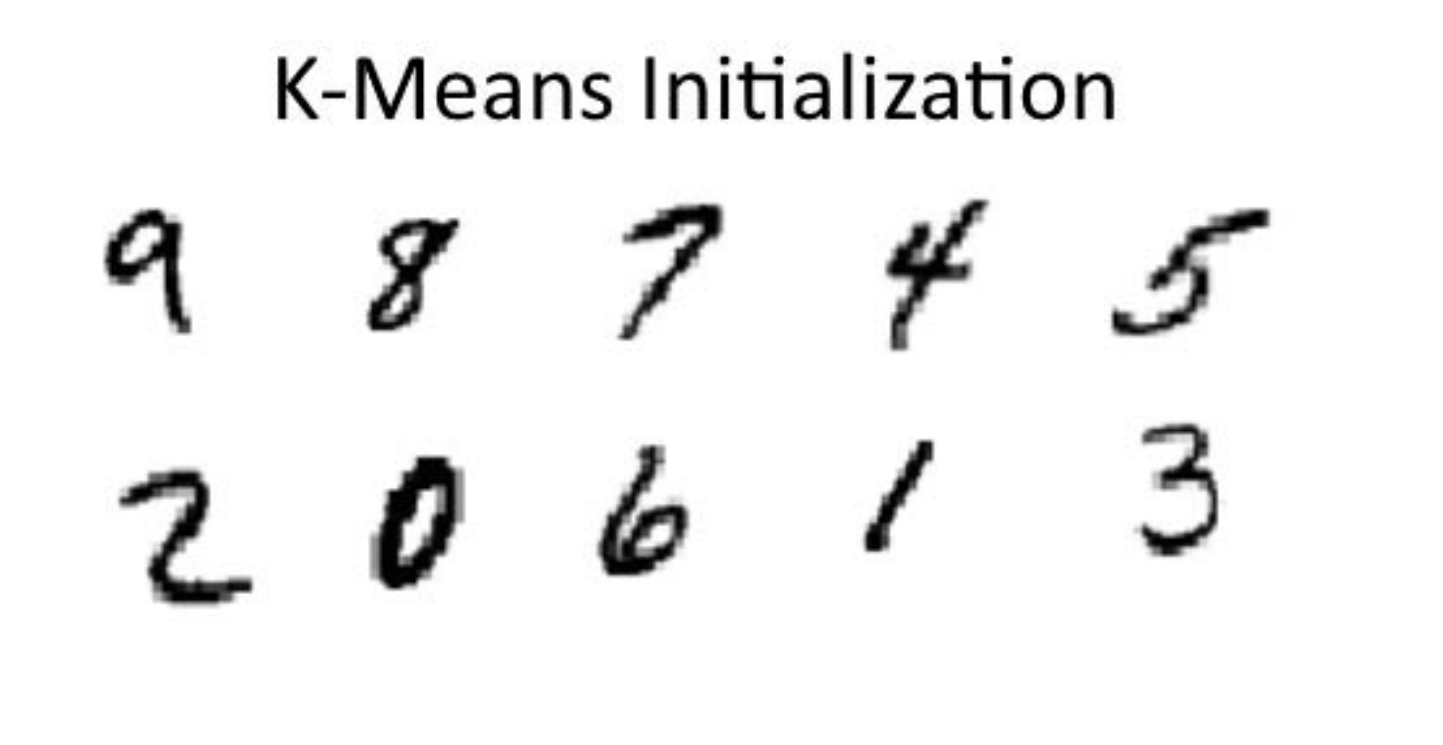}
\caption{Comparison of MNIST unsupervised classification results of 
the LeNet-5 architecture with the random (left) and k-means (right)
initializations, where the images that are closest to centroids
of ten output nodes are shown.}\label{Fig:k-means}
\end{figure}

Once the network is initialized, we can feed the test data to the
network and observe the output, which corresponds to unsupervised
learning. The comparison of unsupervised classification results with the
random and k-means initializations is given in Fig.  \ref{Fig:k-means},
where we show images that are closest to the anchor vectors (or
centroids) of the ten output nodes. We see that the k-means
initialization provides ten anchor vectors pointing to ten different
digits while the random initialization cannot do the same. Different
random initialization schemes will lead to different results, yet the
one given in Fig.  \ref{Fig:k-means} is representative. That is,
multiple anchor vectors will point to the same digit. 

\begin{figure}
\centering
\includegraphics[width=0.4\linewidth]{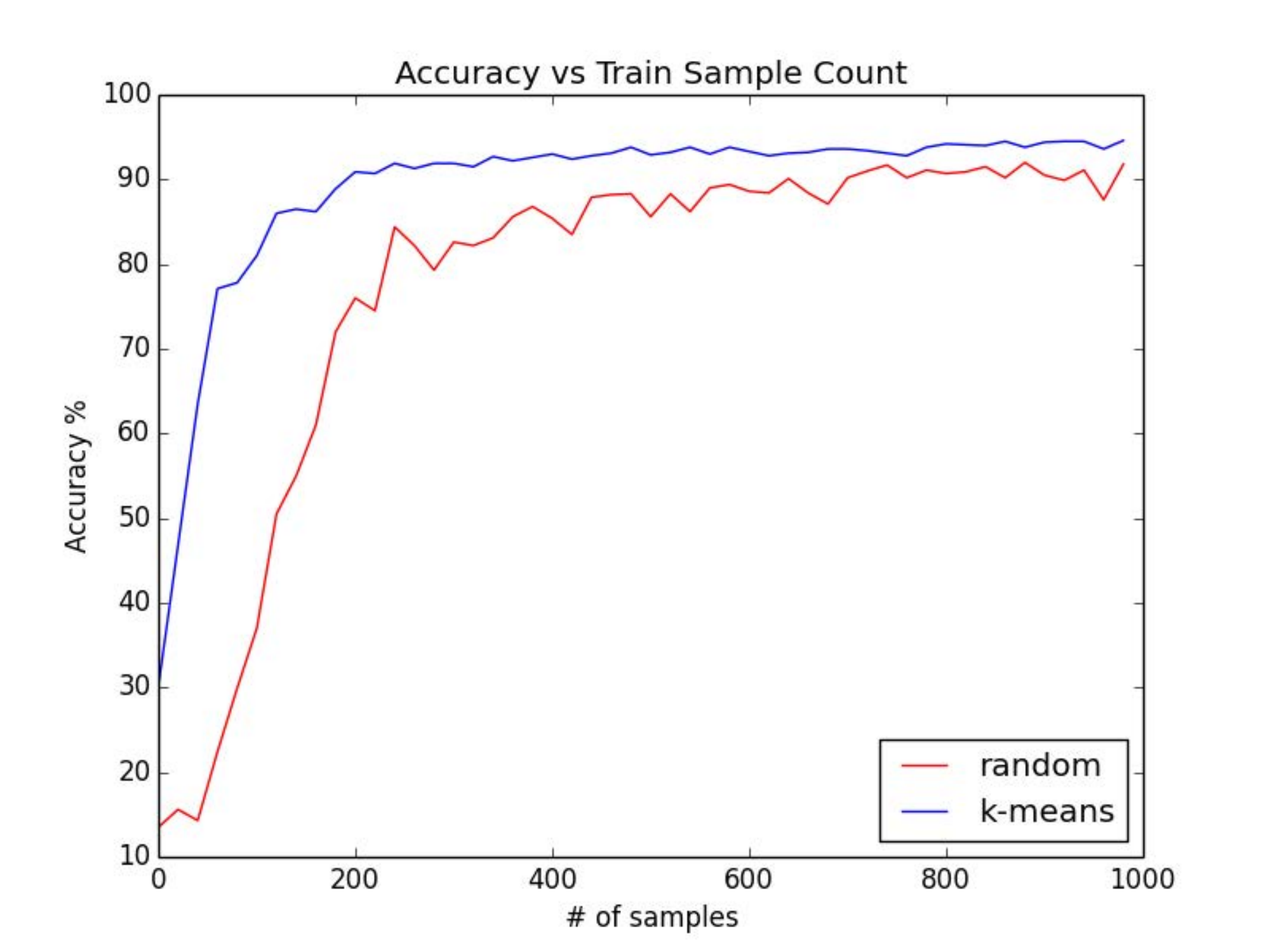}
\caption{Comparison of MNIST weakly-supervised classification results of
the LeNet-5 architecture with the random and k-means initializations, 
where the correct classification rate is plotted as a function
of training sample numbers.}\label{Fig:Learning}
\end{figure}

{\bf Guided Anchor Vector Update.} We apply the BP for network training
with a varying number of training samples. For a fixed number of
training samples, we train the network until its performance converges
and plot the correct classification rate in Fig. \ref{Fig:Learning}. The
two points along the y-axis indicate the correct classification rates
without any training sample. The rates are around 32\% and 14\% for the
k-means and random initializations, respectively. Note that the 14\% is
slightly better than the random guess on the outcome, which is 10\%.
Then, both performance curves increase as the number of training samples
grows.  The k-means can reach a correct classification rate of 90\% when
the number of training samples is around 250, which is only 0.41\% of
the entire MNIST training dataset ({\em i.e.} 60K samples). This shows
the power of the LeNet-5 even under extremely low supervision. 

To further understand the role played by label guidance, we examine the
impact of the BP on the orientation of anchor vectors in various layers.
We show in Table \ref{Table:anchor-vector-change} the averaged
orientation changes of anchor vectors in terms of radian (or degree) for
the two cases in Fig.  \ref{Fig:k-means}. They are obtained after the
convergence of the network with all 60,000 MNIST training samples.  This
orientation change is the result due to label guidance through the BP.
It is clear from the table that, a good network initialization
(corresponding to unsupervised learning) leads to a faster convergence
rate in supervised learning. 

\begin{table}[]
\begin{center}
\resizebox{0.55\textwidth}{!}
{\begin{tabular}{|c|c|c|}  \hline
In/Out layers & k-means & random \\ \hline\hline
Input/S2      &  0.155  (or $8.881^o$) & 1.715 (or $98.262^o$) \\ \hline
S2/S4         &  0.169  (or $9.683^o$) & 1.589 (or $91.043^o$) \\ \hline
S4/C5         &  0.204  (or $11.688^o$) & 1.567 (or $89.783^o$) \\ \hline
C5/F6         &  0.099  (or $5.672^o$) & 1.579 (or $90.470^o$) \\ \hline
F6/Output     &  0.300  (or $17.189^o$) & 1.591 (or $91.158^o$) \\ \hline
\end{tabular}}
\end{center}
\caption{The averaged orientation changes of anchor vectors in terms of
the radian (or degree) for the k-means and the random initialization schemes.}
\label{Table:anchor-vector-change}
\end{table}

\begin{figure}
\centering
\includegraphics[width=0.8\linewidth]{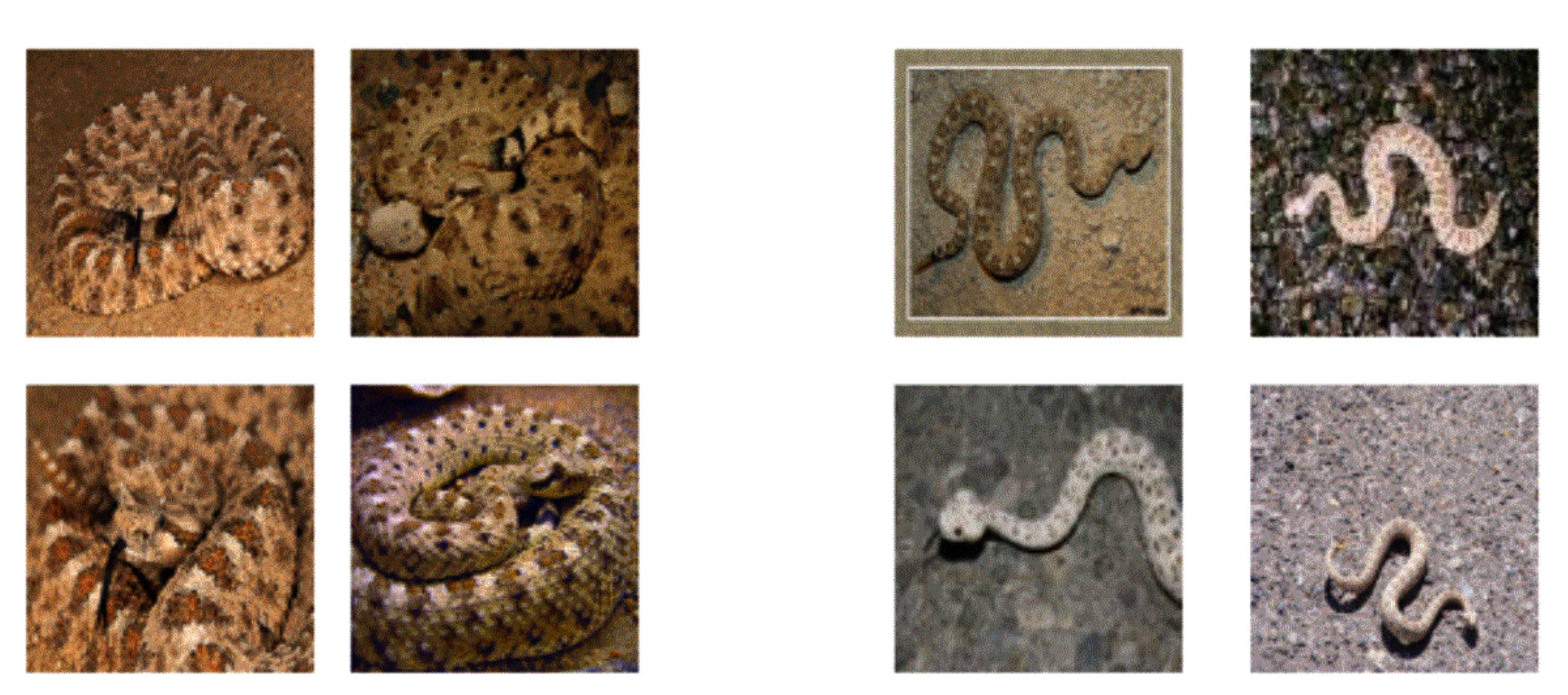}
\caption{Two sub-classes obtained from the horned rattlesnake class using
unsupervised clustering.}\label{Fig:snake}
\end{figure}

{\bf Classes and Sub-Classes.} We use another example to gain further
insights to the guided clustering process. We can zoom into the horned
rattlesnake class obtained by the AlexNet and conduct the unsupervised
k-means on feature vectors in the last layer associated with this class
to further split it into multiple sub-classes.  Images of two
sub-classes are shown in Fig.  \ref{Fig:snake}. Images in the same
sub-classes are visually similar.  However, they are not alike across
sub-classes.  The two sub-classes are grouped together under the horned
rattlesnake class because they share the same class label (despite
strong visual dissimilarity). That shows the power of label guidance.
However, the feature distance is shorter for images in the same
sub-class and longer for images in different sub-classes. This is due to
the inherent clustering capability of CNNs. 

\subsection{Discussion and Open Issues} \label{sec:E}

{\bf Discussion.} A CNN was viewed as a guided multi-layer RECOS
transform in this note.  The following known facts can also be explained
using this interpretation. 
\begin{itemize}
\item {\bf Robustness to wrong labels.} Humans do clustering first and
then the CNN mimics humans based on the statistics of all labeled
samples. It can tolerate small percentages of erroneous labels since
these wrongly labeled data do not have a major impact on clustering
results. 

\item {\bf Overfitting.} Overfitting occurs when a statistic model
describes noise instead of the underlying input/output relationship. For
a given number of observations, this could happen for an excessively
complex model that has too many model parameters.  Such a model has poor
prediction performance since it over reacts to minor fluctuations in the
training data. Although a CNN has a large number of parameters (namely,
filter weights), it does not suffer much from overfitting for the
following reason. When there are only input and output layers without
any hidden layers in between, the CNN solves a linear least-squared
regression problem (where no rectification is needed.) It is well known
the linear regression is robust to noisy data.  When there are hidden
layers, the filter weight determination is a cascaded optimization
problem, which has to be solved iteratively.  In the BP process, we
update the filter weights layer by layer in a backward direction.
Fundamentally, it still attempts to solve a regression problem at each
layer. Although a rectifier conducts rectification on the output, it
does not change the regression nature of MLPs and CNNs. 
\item {\bf Data augmentation.} A low-cost way to generate more samples
is data augmentation. This is feasible since minor perturbations in the
image pixel domain do not change their class types. 
\item {\bf Dataset bias.} A CNN can be biased due to the inherent bias
in the low level representation existing in training samples. Thus, the
performance of a CNN can degrade significantly from one dataset to the
other in the same application domain due to this reason. 
\end{itemize}

{\bf Open Issues.} There are many interesting open problems remaining for 
further exploration.
\begin{itemize}
\item {\bf Network Architecture Design.} It is interesting to be able to
specify the layer number and the filter number per layer for given
applications automatically.
\item {\bf Decoder Network Analysis.} The classification network maps an
image to a label. There are image processing networks that accept an
image as the input and another image as the output. Examples include
super-resolution networks, semantic segmentation networks, etc. These
networks can be decomposed into an encoder-decoder architecture. The
analysis in this note focuses on the encoder part. It is interesting to
generalize the analysis to the decoder part as well. 
\item {\bf Localization and Attention.} Region proposals have been used
in object detection to handle the object localization problem. It is
desired to learn the object location and human visual attention from the
network automatically without the use of proposals. The design and
analysis of networks to achieve this goal is interesting. 
\item {\bf Transfer Learning.} It is often possible to finetune a CNN
for a new application based on an existing CNN model trained by another
dataset in another application. This is because the low-level image
representation corresponding to the beginning CNN layers can be very
flexible and equally powerful. 
\item {\bf Weakly Supervised Learning.} Unsupervised and Heavily
supervised learning are two extremes. Weakly supervised learning occurs
most frequently in our daily applications. The design and analysis of a
weakly supervised learning mechanism based on CNNs is interesting and
practical. 
\end{itemize}
The interpretation of a CNN as a guided multi-layer RECOS transform
should be valuable to the investigation of these topics. 

\section*{\bf CONCLUSION}\label{sec:conclusion}

The operating principle of CNNs was explained as a guided multi-layer
RECOS transform in this note. A couple of illustrative examples were
provided to support this claim.  Several known facts were interpreted
accordingly, and some open issues were pointed out at the end. 

\section*{\bf ACKNOWLEDGEMENT}\label{sec:conclusion}

The author would like to thank the help from Andrew Szot, Shangwen Li,
Zhehang Ding and Gloria Budiman in running experiments and drawing
figures for this work and valuable feedback comments from a couple of
friends, including Bart Kosko, Kyoung Mu Lee, Sun-Yuan Kung and
Jenq-Neng Hwang.  This material is based on research sponsored by DARPA
and Air Force Research Laboratory (AFRL) under agreement number
FA8750-16-2-0173. The U.S. Government is authorized to reproduce and
distribute reprints for Governmental purposes notwithstanding any
copyright notation thereon. The views and conclusions contained herein
are those of the author and should not be interpreted as necessarily
representing the official policies or endorsements, either expressed or
implied, of DARPA and Air Force Research Laboratory (AFRL) or the U.S.
Government. 

\section*{\bf AUTHOR}\label{sec:authors}

C.-C. Jay Kuo (cckuo@ee.usc.edu) is a Professor of Electrical
Engineering at the University of Southern California, Los Angeles,
California, USA. 

\bibliographystyle{IEEEtran}
\bibliography{CNN}


\end{document}